\def\year{2021}\relax
\documentclass[letterpaper]{article} 
\usepackage{aaai21}  
\usepackage{times}  
\usepackage{helvet} 
\usepackage{courier}  
\usepackage[hyphens]{url}  
\usepackage{graphicx} 
\usepackage{algorithm}
\usepackage{algorithmic}
\usepackage{latexsym}
\usepackage{url}
\usepackage{booktabs}
\usepackage{multirow}
\usepackage{natbib}  
\usepackage{caption} 
\title{WeChat AI \& ICT's Submission for DSTC9 Interactive Dialogue Evaluation Track}
\author{
    \textbf{Zekang Li$^{124}$, Zongjia Li$^{23}$, Jinchao Zhang$^2$, Yang Feng$^1$\thanks{Joint work with Pattern Recognition Center, WeChat AI, Tencent Inc, China. Yang Feng is the corresponding author. This work was done when Zekang Li and Zongjia Li were interning at Pattern Recognition Center, WeChat AI, Tencent.}, Jie Zhou$^2$}\\
\textsuperscript{\rm 1}Key Laboratory of Intelligent Information Processing\\ 
Institute of Computing Technology, Chinese Academy of Sciences \\
\textsuperscript{\rm 2}WeChat AI, Tencent Inc, China\\
\textsuperscript{\rm 3}School of EECS, Peking University \\
\textsuperscript{\rm 4}University of Chinese Academy of Sciences \\
{\tt \small \{lizekang19g, fengyang\}@ict.ac.cn, zongjiali@pku.edu.cn}\\
{\tt \small \{dayerzhang, withtomzhou\}@tencent.com}
}

\begin{document}

\maketitle

\begin{abstract}
We participate in the DSTC9 Interactive Dialogue Evaluation Track \cite{gunasekara2020overview} sub-task 1 (Knowledge Grounded Dialogue) and sub-task 2 (Interactive Dialogue). In sub-task 1, we employ a pre-trained language model to generate topic-related responses and propose a response ensemble method for response selection. In sub-task2, we propose a novel Dialogue Planning Model (DPM) to capture conversation flow in the interaction with humans. We also design an integrated open-domain dialogue system containing pre-process, dialogue model, scoring model, and post-process, which can generate fluent, coherent, consistent, and human-like responses. We tie 1$^{st}$ on human ratings and also get the highest Meteor, and Bert-score in sub-task 1, and rank 3$^{rd}$ on interactive human evaluation in sub-task 2.
\end{abstract}

\section{Introduction}
Our WeChat AI team participates in the DSTC9 Interactive Dialogue Evaluation Track sub-task 1 and sub-task 2. Sub-task 1 is knowledge-grounded dialogue generation on the Topical-Chat dataset, which is evaluated in a static manner. Sub-task 2 aims to extend dialog models beyond datasets and interactively evaluates dialogue systems with real users on DialPort \cite{zhao2016dialport}. We mainly focus on improving the topical relevance of responses in sub-task 1 and improving topic depth, consistency, human-likeness in sub-task 2. 

In the sub-task 1 (Knowledge Grounded Dialogue), our model architecture is built on the GPT2 model \cite{radford2019language}. We mainly focus on exploring better decoding methods and ensemble methods. In the decoding stage, sampling-based methods are likely to generate more diverse and human-like responses compared to beam-search. But sampling-based methods always suffer from less topical relevance. To improve the topical relevance of responses, we propose a metric-based ensemble method for response selection. 

In the sub-task 2 (Interactive Dialogue), the ultimate goal of the open domain dialogue system is to interact with real users effectively. Recently, due to pre-training with large scale transformer models, open-domain dialogue systems \cite{zhang2019dialogpt,bao2020plato,smith2020can,adiwardana2020towards} have achieved great success. However, based on our real interaction experience with the systems, they can be improved on flexibility, topic depth, and consistency. 

As to flexibility, current dialogue models are often lost in the dialogue when real users change the topic by accident because there is little topic change in the training data. Therefore, we propose a simple but efficient data augmentation method by constructing more flexible training data. 

As to topic depth, as shown in Figure \ref{fig:DPM}, we consider the whole dialogue process as many vector operations and propose the Dialogue Planning Model to capture the topic flow in the dialogue, which can improve topic depth effectively in the interaction with real users. 

As to consistency, many dialogue models are likely to ignore the information in history when generating responses, such as opinion conflicts, personal information conflicts, and so on, which severely harms the dialogue consistency. To improve dialogue consistency, we employ a natural language inference model to detect the responses that conflict with dialogue history.

\begin{figure*}[t]
    \centering
    \includegraphics[width=0.9 \textwidth]{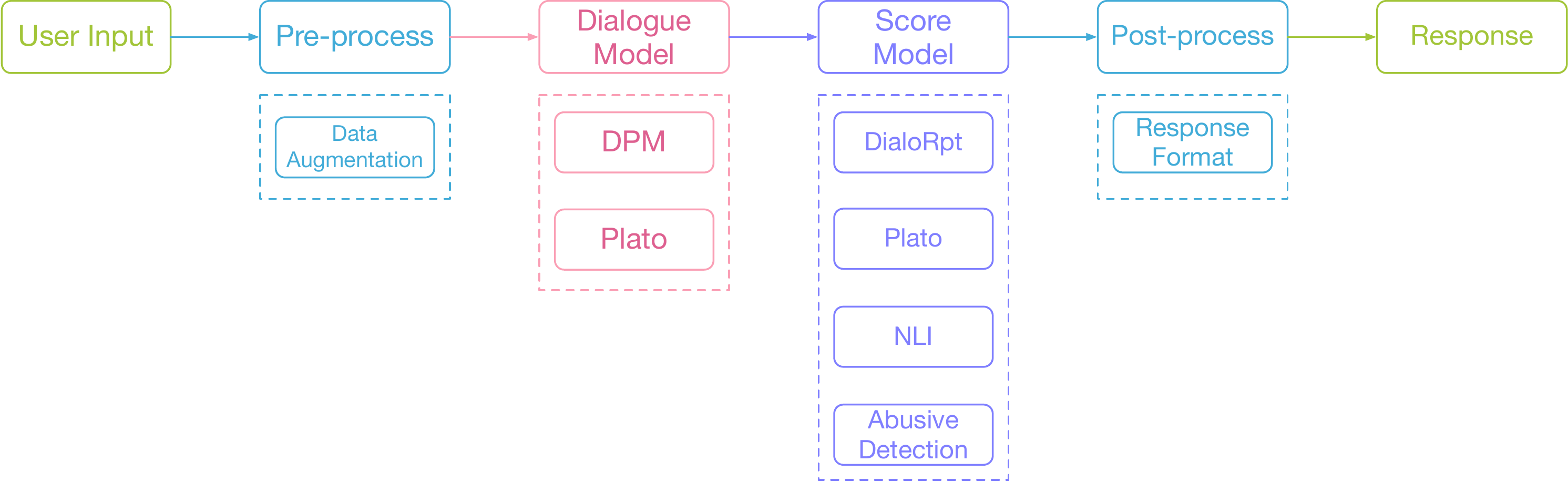}
    \caption{Overview of the interactive dialogue system. It contains 4 modules: Pre-process, Dialogue Model, Score Model, and Post-process. Pre-process mainly contains data augmentation. Dialogue Model includes two dialogue generation model: DPM and Plato, and generates response candidates. Score Model aims to select the most appropriate response through overall score by four scoring model. Post-process is used to make the response more human-like.}
    \label{fig:system}
\end{figure*}

\section{Related Work}
\subsection{Knowledge-Grounded Open-Domain Dialogue.}
Knowledge-grounded open-domain dialogue is an important step towards a human-like dialogue system. Recent works mainly obtain knowledge from knowledge graphs \cite{zhou2018commonsense,tuan2019dykgchat}, from unstructured text \cite{dinan2018wizard,li2019incremental,lian2019learning,kim2020sequential}, and from visual information \cite{li2020bridging,huber2018emotional}. In the sub-task 1, we focus on using unstructured text as knowledge. Rather than learning from scratch like most recent work, we utilize a pre-trained language model and propose a response ensemble method to generate responses with more knowledge relevance.

\subsection{Pre-trained Language Model.}
Pre-trained Language models have brought about much improvement on various NLP tasks. GPT2 \cite{radford2019language} and BERT \cite{devlin2018bert} are representative uni-directional and bi-directional language models. Based on GPT2, DialoGPT \cite{zhang2019dialogpt} is trained for dialogue response generation using Reddit comments. Meena \cite{adiwardana2020towards} utilize more social media data to make the chatbot more human-like. Besides, Blender fine-tunes the pre-trained model on human-annotated conversations. Furthermore, towards more diverse, human-like responses, Plato \cite{bao2020plato} introduces discrete latent variable and curriculum learning in the training process. To improve dialogue coherence and topic depth, we introduce a dialogue flow method upon the GPT2 model.

\section{Our Method}
For sub-task 1 (Knowledge Grounded Dialogue), we fine-tune the large pre-trained language model on the Topical-Chat dataset to generate topic-related knowledge grounded response candidates. Besides, we propose a response ensemble method to improve the topical relevance of response.
For sub-task 2 (Interactive Dialogue), we build a dialogue system, which consists of four modules: pre-process, dialogue model, score model, and post-process, as shown in Figure \ref{fig:system}.

\begin{figure*}[t]
    \centering
    \includegraphics[width=0.8 \textwidth]{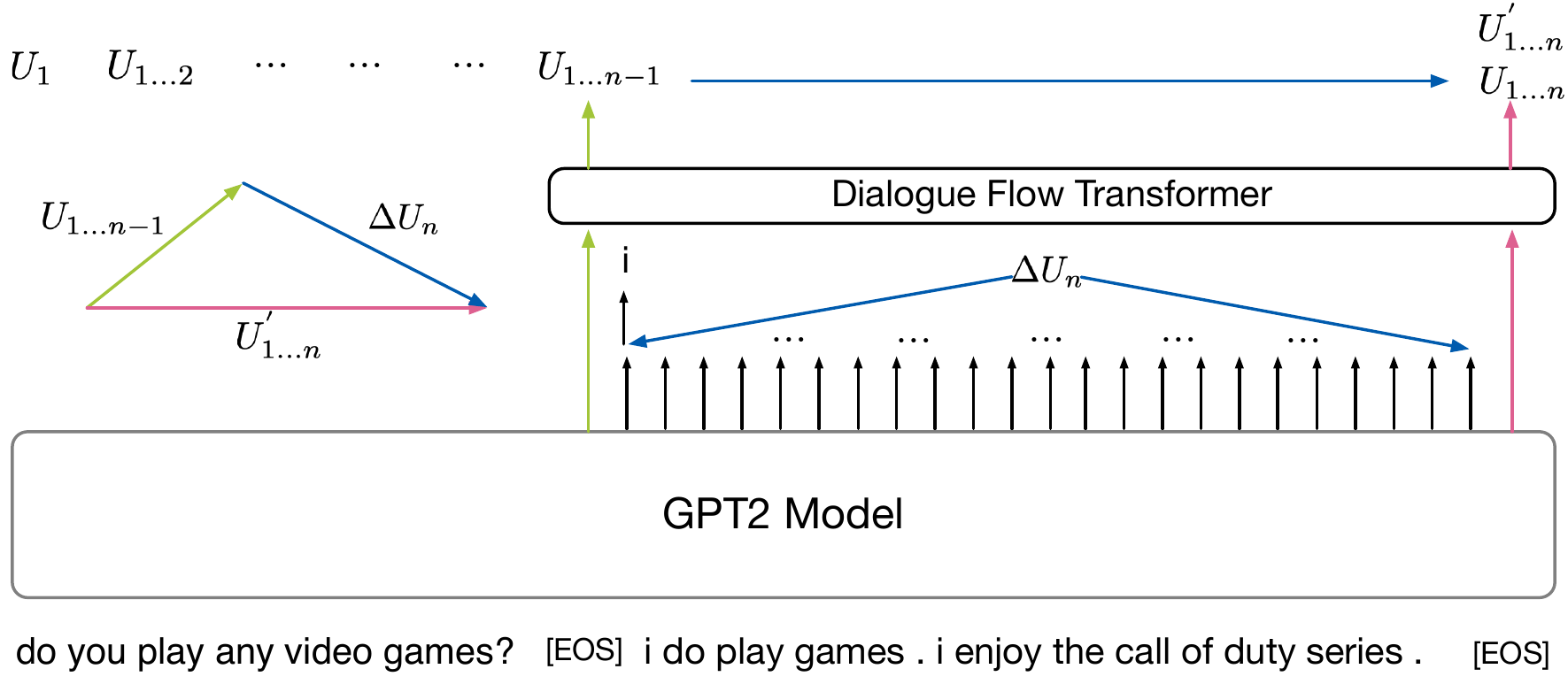}
    \caption{Dialogue planning model (DPM). We consider the whole dialogue process as many vector operations. $U_{1...n}$ denotes the representation of $1\sim n$ utterance encoded by GPT2 model. The representation of utterance $u_n$ conditioned on $u_{\textless n}$ can be calculated by $U_{1...n}^{'} - U_{1...n-1}$, where $U_{1...n}^{'}$ is the predicted representation by the dialogue flow transformer block.}
    \label{fig:DPM}
\end{figure*}
\subsection{Sub-task 1: Knowledge-Grounded Dialogue Generation}
This task is to generate a response to a fixed dialogue context given the topic-related facts. Formally, let $C$ and $R$ denote the dialogue context and response respectively, and $K $ denote the topical-related facts. The probability to generate the response can be computed as:
\begin{equation}
	P(R|C, K; \theta) = \prod_{i=1}^{N}P(R_{i}|C, K, R_{\textless i}; \theta)
\end{equation}
where $\theta$ is the learnable parameter.

\noindent\textbf{Model.} The model architecture is based on gpt2-large \cite{radford2019language}. For fine-tuning, we concatenate the fact $K$, the dialogue context $C$, and the golden response $R$ as the input sequence, and optimize the model by minimizing the following loss:
\begin{equation}
	\mathcal L = - \sum_{i=1}^N log(P(R_{i}|C, K, R_{\textless i}; \theta))
\end{equation}
To generate more diverse and human-like responses, we employ top-p sampling method \cite{holtzman2019curious} rather than greedy decoding and beam search. 

\begin{table*}[tb]
		\centering
		\begin{tabular}{r@{~} p{10cm} c@{~~~}c}
			\toprule[1pt]
            \bf  & \bf Sampled Text & \bf Meteor-self & \bf Meteor-GT \\ 
			\hline
			\bf{Example 1:} & the warriors played the nba finals at the cow palace because the oakland arena was booked. &  0.652  & 0.592 \\
			\bf{Example 2:} & the golden state warriors played the home games in the 1975 nba finals at the cow palace. & 0.590 & 0.591 \\
			\bf{Example 3:} & the cow palace was the place to watch games in 1975. &    0.341 & 0.540\\
			\bf{Example 4:} & the golden state warriors played at the cow palace because the oakland arena was booked. & 0.657 & 0.808 \\
			\bf{Example 5:} & the golden state warriors played in 1975 at the cow palace because the oakland arena was booked. & \bf 0.734 & \bf 0.811\\
			\bf{Ground-truth:} & in 1975 the golden state warriors had to play at the cow palace because their arena was booked. & - & -\\
			\bottomrule[1pt]
		\end{tabular}
		\caption{Response examples for the response ensemble method. As shown in the table, the sampled texts always lack of different parts of information. The response ensemble method can select the most integrated response. Meteor-self means the Meteor score with the other texts and Meteor-GT means the Meteor score with the ground-truth.}
		\label{tab:example}
	\end{table*}

\begin{algorithm}
\caption{Metric-based ensemble method for response selection.}
\label{alg:ensemble}
\begin{algorithmic}[1]
\STATE \textbf{Input: } Response candidates $\{r_i| i=1,2,...,N\}$.\
\STATE \textbf{Output: } The most topic-related response $r$.\
\STATE Select a metric, such as Meteor\, Bert-score, and BLEU;\
\STATE Initialize metric score $M = \{m_i=0|i=1,2,...,N\}$;\
\FOR{each $i \in [1,N]$}
\FOR{each $j \in [1,i) \cup (i, N]$}
\STATE $m_i = m_i + Metric(r_i, r_j)$\
\ENDFOR
\STATE $m_i = m_i / (N-1)$\
\ENDFOR
\STATE \textbf{Return:} $r_{argmax(M)}$.
\end{algorithmic}
\end{algorithm}

\noindent\textbf{Response Ensemble Method.}
Sampling-based decoding methods always bring about more diversity but suffer from less topical relevance. To improve the topical relevance, we propose a metric-based ensemble method to select the most topical-relevant response from the generated response candidates, as shown in Algorithm \ref{alg:ensemble}. The method can improve the information integrity as well as the reference-based metric performance. As shown in Table \ref{tab:example}, the 5 sampled texts lack of different parts of information. Through the method, we can select the most integrated and appropriate response.

\subsection{Sub-task 2: Interactive Dialogue}
There are many challenges to building an effective dialogue system, such as improving dialogue consistency, dialogue topic depth, and human-likeness. In this part, we will describe our methods to build and improve the dialogue system. \\
\noindent\textbf{System Overview.}
As shown in Figure \ref{fig:system}, our interactive open-domain dialogue system contains four modules: Pre-process, Dialogue Model, Score Model, and Post-process.

\noindent\textbf{Pre-process.} During the interaction, real users always change the dialogue topic by accident and have different preferences on topic depth. Therefore, we propose a simple but efficient data augmentation method to handle these problems. For different topic depth, given dialogue $A$, we randomly cut out some utterances from the end and get a new dialogue $C$. For the topic changes, given dialogue $C$ and dialogue $B$, we concatenate $C$ and $B$ as a new dialogue $D$. In the training stage, we randomly sample dialogue $C$ and $D$ according to a fixed probability.

\noindent\textbf{Dialogue Model.}
To generate diverse, informative, fluent responses, in the Dialogue model module, we employ two models: Dialogue Planning Model (DPM) and Plato \cite{bao2020plato} which is a large-scale pre-trained dialogue generation model with discrete latent variable. These two models generate many response candidates for the following scoring modules \footnote{We use the public released Plato model (\url{https://github.com/PaddlePaddle/Knover/tree/master/plato-2}).}. 

\begin{table*}[tb]
		\centering
		\begin{tabular}{l@{~~~~} c@{~~~~}c@{~~~~}c@{~~~~} c@{~~~~}c}
			\toprule[1pt]
            \bf Models & \bf Meteor & \bf Bert-score & \bf USR & \bf Human Rating \\ 
			\hline
			\bf Our model  &  0.142 ~&  0.869 ~& - ~& -  \\
			~~~ \bf + bert-score ensemble &  0.147 ~& \bf ~~0.876$^*$  ~& 4.34 ~& 4.13  \\
			~~~ \bf + meteor ensemble &  \bf ~~0.160$^*$ ~& 0.873  ~& \bf 4.51 ~& 4.21 \\
			\bf Sub-task 2 system &  0.070 ~& 0.843  ~& 3.86 ~&  \bf ~~4.28$^*$  \\
			\bottomrule[1pt]
		\end{tabular}
		\caption{Automatic and human evaluation results on the test set provided by the organizers in DSTC9 Interactive Dialogue Evaluation Track sub-task 1. Note that $^*$ denotes that it ranks 1$^{st}$ over all submissions in the competition.}
		\label{tab:static}
	\end{table*}
	
\begin{table}[tb]
		\centering
		\begin{tabular}{l@{~~~~} c@{~~~~}c}
			\toprule[1pt]
            \bf Models & \bf FED & \bf Human Rating \\ 
            \hline
            \multicolumn{3}{c}{\it Baseline} \\
			\hline
            \bf Transformer & 3.69 ~& 3.60 \\ 
            \bf DialoGPT & \bf 4.72 ~& 3.87 \\
            \hline
            \multicolumn{3}{c}{\it Our system} \\
			\hline
			\bf Our System  &  4.61 ~& \bf 4.08  \\
			\hline
		\end{tabular}
		\caption{Automatic and human evaluation results on interactive dialogues provided by the organizers in DSTC9 Interactive Dialogue Evaluation Track sub-task 2.}
		\label{tab:interactive}
	\end{table}

\noindent\textbf{Dialogue Planning Model (DPM).} 
As shown in Figure \ref{fig:DPM}, to improve the dialogue coherence and topic depth, we design the Dialogue Planning Model based on the gpt2 model. Particularly, we design a dialogue flow transformer block (FLOW) upon the GPT2 model. Formally, given a dialogue containing $n$ utterances $u = [u_1, u_2,...,u_n]$, suppose $U_{1...n}$ denote the representation of $1\sim n$ utterance encoded by GPT2 model. We consider dialogue process as many vector operations as shown in Figure \ref{fig:DPM}.  \\
\begin{equation}
	\Delta U_n = U_{1...n}^{'} - U_{1...n-1}
\end{equation}
where $\Delta U_n$ can be considered as the representation of utterance $u_n$ conditioned on $u_{\textless n}$ and $U_{1...n}^{'}$ is the predicted representation by the dialogue flow transformer block.

To train the Dialogue Planning Model, we design three tasks: Dialogue Flow Prediction, Response Generation, Bag-of-Words Prediction. 

Dialogue Flow Prediction is to predict the representation of $1\sim n$ utterances $U_{1...n}^{'}$ based on $U_1, U_{1...2}, ..., U_{1...n-1}$. 
\begin{equation}
	U^{'}_{1...n} = FLOW(U_1, U_{1...2}, ..., U_{1...n-1})
\end{equation}
We train the Dialogue Flow Prediction task by minimizing Mean Squared Error:
\begin{equation}
	\mathcal L_{flow} = MSELoss(U_{1...n}, U_{1...n}^{'})
\end{equation}
Response Generation is to generate dialogue response using utterances $u_{\textless n}$ and predicted representation of utterance $u_n$. Specifically, when generating each token, we concatenate $\Delta U$ and the gpt2 output hidden states. We optimize Response Generation task by minimizing the following loss:
\begin{equation}
	\mathcal L_{gen} = - \sum_{i=1}^N log(P(u_n^i|u_{\textless n}, u_n^{\textless i}, \Delta U_n; \theta))
\end{equation}
Bag-of-Words Prediction task is to predict the words in an utterance using $\Delta U_n$, which can be considered as a topical constraint. This task can be optimized by minimizing the following loss:
\begin{equation}
	\mathcal L_{bow} = -\sum_{i=1}^N log(u_{n}^i|\Delta U_n)
\end{equation}
The overall loss to train Dialogue Planning Model can be computed as follows:
\begin{equation}
	\mathcal L = \mathcal L_{flow} + \mathcal L_{gen} + \mathcal L_{bow}
\end{equation}

\noindent\textbf{Score Model.} Score Model consists of four scoring models: DialoRpt, Plato, NLI, Abusive Detection. DialoRPT \cite{gao2020dialogrpt} is a large-scale dialog ranking model based on the human feedback of dialogue responses. Plato denotes the response selection model in \citet{bao2020plato}.  NLI means natural language inference, which is useful for non-consistency detection. We exploit RoBERTa-large-mnli \cite{liu2019roberta} to predict whether the response conflicts with dialogue history. Abusive Detection is to detect some abusive words. 

Plato always provides relatively close scores (e.g. 1e-4) for top 10 responses, while DialoRPT gives scores with a larger gap. In our interactive experiments, DialoRPT always choose some responses that are not so coherent with the context, which is may be due to that it is trained based on the human likert-score in the forum. Therefore, we first use remove the responses with abusive words, and exploit plato to filter relatively coherent response candidates and then use NLI and DialoRPT to choose the best one. The whole score process is shown in Algorithm \ref{alg:score}. 

\begin{algorithm}
\caption{Scoring method for interactive dialogue response selection.}
\label{alg:score}
\begin{algorithmic}[1]
\STATE \textbf{Input: } Response candidates $R = \{r_i| i=1,2,...,N\}$, dialogue history.\
\STATE \textbf{Output: } The most appropriate response $r$.\
\STATE Detect abusive words and remove the response with abusive words.
\STATE Score by Plato and select top-10 responses.
\STATE For each response, use the NLI model to detect conflict with dialogue history. Remove the response with conflict.
\STATE Score by DialoRPT and select top-1 response.
\STATE \textbf{Return:} $r_{argmax(M)}$.
\end{algorithmic}
\end{algorithm}

\noindent\textbf{Post-process.} To respond with more human-like responses, after selecting the most appropriate response through the scoring model, post-process mainly formats the response more human-like, such as uppercase problem on special entities.

\section{Experiments and Analysis}
\subsection{Experiment Settings}
For sub-task 1, we conduct experiments on the Topical-chat dataset \cite{gopalakrishnan2019topical}, which contains dialogues with topical knowledge. The dataset contains 9058 dialogues for training, 565 dialogues for freq validation, and 565 dialogues for freq test. We only train our model on the training data without any other data and select the model based on the performance on the freq validation data. We employ 60 response candidates for the response ensemble. 

For sub-task 2, we initialize the Dialogue Planning Model with GPT2-large\footnote{https://github.com/huggingface/transformers} and fine tune it on the BST dataset, which contains four human annotated conversations datasets: ConvAI2  \cite{zhang2018personalizing,dinan2020second}, Empathetic Dialogues \cite{rashkin2018towards}, Wizard of Wikipedia \cite{dinan2018wizard}, Blended Skill Talk \cite{smith2020can}. We train the model on 8 Tesla V100 GPUs for about 24 hours.

\subsection{Metrics}
For sub-task 1, we employ the following metrics to evaluate our model.\\
\noindent\textbf{Bert-score:} A reference-based evaluation metric that uses a pre-trained BERT model to greedily match each word in the generated response with the ground-truth response \cite{zhang2019bertscore}.  \\
\noindent\textbf{Meteor:} A reference-based evaluation metric which is designed as an improvement on BLEU \cite{papineni2002bleu} using a harmonic mean of precision and recall \cite{banerjee2005meteor}.\\
\noindent\textbf{USR:} An unsupervised and reference-free evaluation metric for response evaluation \cite{mehri2020usr}. \\
\noindent\textbf{Human Ratings:} Human evaluation is carried out on Amazon Mechanical Turk with the annotation questionnaire used in FED score \cite{mehri2020unsupervised}. There are 100 context-response pairs sampled and each one is labeled by 3 annotators. 

For sub-task 2, we submit our system on DialPort and collect dialogs through conversations with real users. The organizers mainly use human evaluation as well as the FED score to evaluate our dialogue system. \\
\noindent\textbf{FED:} A reference-free evaluation metric \cite{mehri2020unsupervised} which is designed to evaluate the interactive dialogue with real users.\\
\noindent\textbf{Human Ratings:} The human evaluation is carried out on Amazon Mechanical Turk with the annotation questionnaire used in the FED score \cite{mehri2020unsupervised}. There are 200 dialogs evaluated for each system. \\

\subsection{Experiment Results}
\noindent\textbf{Sub-task 1: } The evaluation results on test set for sub-task 1 is shown in Table \ref{tab:static}. Our model with bert-score ensemble reaches the best bert-score  and with meteor ensemble gets the best meteor score over all submissions. We also use the interactive system in sub-task 2 for this task, which tie 1$^st$ on human ratings.

\noindent\textbf{Sub-task 2:} As shown in Table \ref{tab:interactive}, on human ratings, our system significantly outperforms baseline system Transformer and DialoGPT, which is provided by the organizer. However, on the FED score, our system is a little lower than DialoGPT. We consider that the FED score is based on DialoGPT, so the DialoGPT system gets a better FED score.

\section{Conclusion}
In this paper, we introduce the WeChat AI's submission for DSTC9 Interactive Dialogue Evaluation Track sub-task 1 and sub-task 2. In sub-task 1, our model is based on GPT2 and we propose a simple but efficient ensemble method for knowledge-grounded dialogue. Our method achieved the highest Bert-score, Meteor, and human ratings using different systems in the competition.  In sub-task 2, to improve topic depth and dialogue coherence, we propose the Dialogue Flow Model and we build an integrated open-domain dialogue system containing four modules: pre-process, dialogue model, scoring model, and post-process for generating more human-like responses. Our system significantly outperforms the baseline methods and ranks 3$^rd$ over all submissions.

\section{Acknowledgement}
We sincerely thank the anonymous reviewers for their thorough reviewing and valuable suggestions. This work is supported by National Key R\&D Program of China (NO. 2018YFC0825201 and NO. 2017YFE0192900).

\bibliography{aaai21}
\end{document}